\documentclass[times,twocolumn,final]{elsarticle}

\usepackage{ycviu0}
\usepackage{framed,multirow}

\usepackage{amssymb}
\usepackage{latexsym}
\usepackage{color}
\usepackage{float}
\usepackage{booktabs}
\usepackage{amsmath}
\usepackage{graphicx}
\usepackage{url}
\usepackage{xcolor}
\definecolor{newcolor}{rgb}{.8,.349,.1}
\newcommand{\etal}{\emph{et al.}}
\newcommand{\eg}{\emph{e.g.}}

\newfloat{algorithm}{tbp}{loa}
\providecommand{\algorithmname}{Algorithm}
\floatname{algorithm}{\protect\algorithmname}

\begin{document}

\begin{frontmatter}

\title{Learning the Compositional Spaces for Generalized Zero-shot Learning}

\author[1]{Hanze {Dong}} 

\author[1]{Yanwei Fu\corref{cor1}}
\cortext[cor1]{Corresponding author}
\author[4]{Sung~Ju~Hwang}
\author[5]{Leonid~Sigal  }
\author[2]{Xiangyang~Xue }
\address[1]{ School of Data Science, Fudan University, Shanghai, China}
\address[2]{School of Computer Science, Fudan University, Shanghai, China}
 
 \address[4]{ KAIST, Seoul, South Korea}
 \address[5]{  University of British Columbia, Vancouver, Canada}
\received{1 May 2013}
\finalform{10 May 2013}
\accepted{13 May 2013}
\availableonline{15 May 2013}
\communicated{S. Sarkar}

\begin{abstract}
This paper studies the problem of Generalized Zero-shot Learning (G-ZSL),
whose goal is to classify instances belonging to both seen and unseen
classes at the test time. We propose a novel space decomposition method
to solve G-ZSL. Some previous models with space decomposition operations
only calibrate the confident prediction of source classes (W-SVM \cite{scheirer2014probability})
or take target-class instances as outliers \cite{socher2013zero}.
In contrast, we propose to directly estimate and fine-tune the decision
boundary between the source and the target classes. Specifically,
we put forward a framework that enables to learn compositional spaces
by splitting the instances into \emph{Source}, \emph{Target}, and
\emph{Uncertain} spaces and perform recognition in each space, where
the uncertain space contains instances whose labels cannot be confidently
predicted. We use two statistical tools, \emph{namely,} bootstrapping
and Kolmogorov-Smirnov (K-S) Test, to learn the compositional spaces
for G-ZSL. We validate our method extensively on multiple G-ZSL benchmarks,
on which it achieves state-of-the-art performances. The codes are available on \url{ https://github.com/hendrydong/demo_zsl_domain_division}.
\end{abstract}


\end{frontmatter}



\section{Introduction}

Zero-shot learning (ZSL) has attracted extensive attention from various
research areas of computer vision. It aims at recognizing novel \emph{target}
classes that are unseen at the training stage by transferring knowledge
from observed\emph{ source} (or auxiliary) categories with many labeled
instances. To enable the knowledge transfer, semantic representations,
auxiliary information such as visual attributes \cite{lampert2014attribute}
and word embeddings \cite{transductiveEmbeddingJournal} are used
to relate target classes with source classes. Typically, most approaches
formulate ZSL as a visual-semantic alignment problem: an \emph{embedding}
space is learned on source classes by transforming their instances
from the visual to semantic space \cite{DeviseNIPS13,fu2016semi,lampert2014attribute,norouzi2013zero},
or vice versa \cite{changpinyo2016synthesized,romera2015embarrassingly,kodirov2017semantic};
in the learned embedding space, such a transformation is applied to
project unseen data onto the space for classification.

In the general experimental setting of ZSL, test instances only come
from target unseen classes. However, this is an unrealistic simplification
of the object categorization tasks \emph{in the wild}. In consequence,
a more realistic setting -- Generalized Zero-Shot Learning (G-ZSL)
where test instances come from both source and target classes, is
considered as a more realistic benchmark of ZSL performance \cite{wild_0shot,verma2018generalized,felix2018multi}.

\begin{figure*}[t]
	\begin{centering}
		\begin{tabular}{cc}
			\includegraphics[width=1\columnwidth]{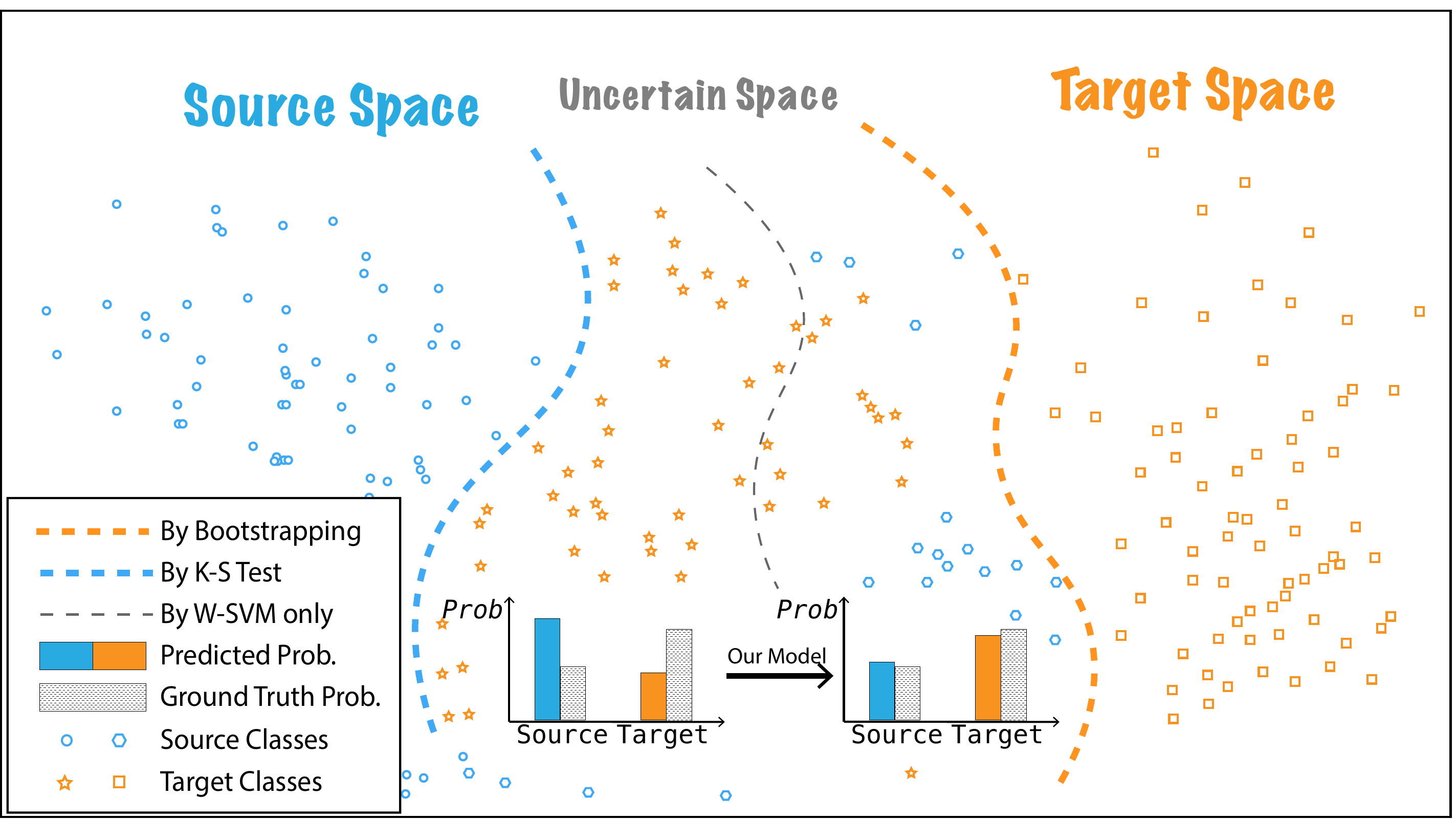}  & \includegraphics[width=1\columnwidth]{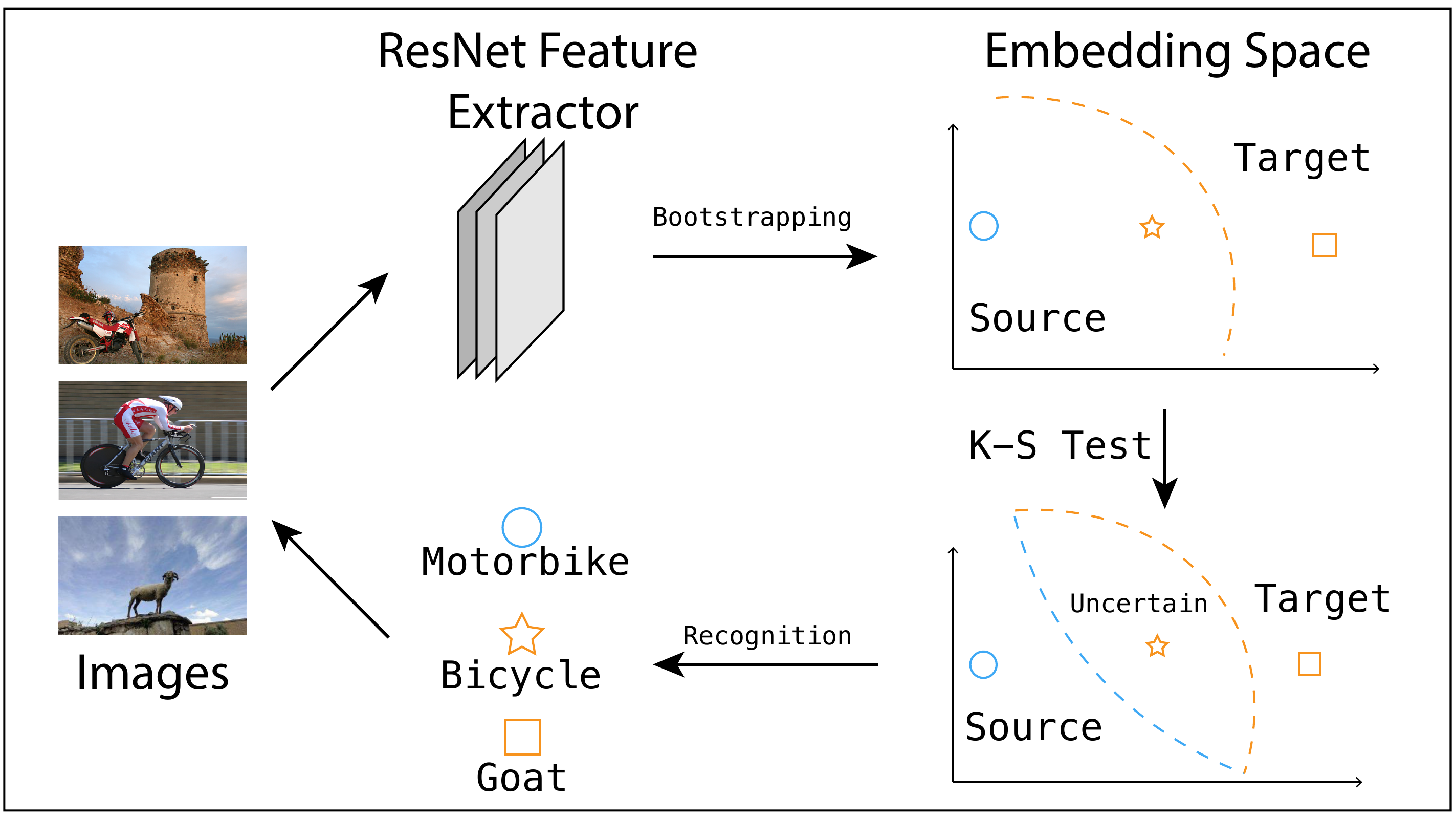}\tabularnewline
			(a) Visual/semantic embedding space  & (b) Our pipeline \tabularnewline
		\end{tabular}
		\par\end{centering}
	\centering{}\caption{\label{fig:Domain-Division} (a) The decision boundary in visual/semantic
		embedding space. We introduce a uncertain space to contain the potentially
		under-confident source instances; (b) The pipeline of our model. The
		initial boundary is estimated by bootstrapping. We can further divide
		a uncertain space by K-S Test. Then, we can recognize instances in each
		compositional space. (Orange: Instances from target classes; Blue:
		Instances from source classes. Different shapes indicate various classes.)}
\end{figure*}

The G-ZSL is still addressed in the form of learning a visual-semantic
alignment with an assumption that the distributions of classes in
the semantic and visual spaces are relatively similar \cite{felix2018multi}.
In contrast to ZSL, approaches of G-ZSL are prone to be biased toward
target classes, resulting in poor classification accuracy, especially
for the target classes \cite{wild_0shot,xian2017zero}. As demonstrated
in Fig. \ref{fig:Domain-Division} (a), the decision boundary between
source and target classes will be inevitably shifted to source classes
in the learned visual embedding space, as target data are unavailable
during the optimization stage of these approaches; thus, a large portion
of target data will be misclassified as source classes at the test
phase, which can also be recognized as a sort of overfitting.

To alleviate such a bias, Long \emph{et al}. \cite{long2017tpami}
optimized the semantic spaces to learn a better transformation from
visual to semantic space. Besides, Generative Adversarial Networks
(GANs) based approaches \cite{verma2018generalized,xian2018feature}
synthesized visual representations of target classes. Liu \emph{et
	al.} \cite{deep_calibration_gzsl} calibrated both confidence of source
classes and uncertainty of target classes. Notably, most of these
previous works are built upon the assumption that the distributions
of classes should be similar in the semantic and visual spaces.

Fundamentally, we argue that \emph{source and target classes }may
follow largely disparate visual/semantic distributions. Therefore,
it is unnecessary to predict source classes in the semantic space.
Such an intuition can be incorporated in addressing the prediction
bias in G-ZSL. It is thus surprising to note that there is little
if any existing work that fully and explicitly explores different
distributions between source and the target classes. \emph{Is it because
	there is a trivial extension of encoding such a new idea?} By training
a supervised mapping from the visual and semantic space, it is indeed
possible to implicitly take the G-ZSL as an outlier detection task
\cite{RichardNIPS13}: determining whether a given test instance is
on the manifold of source classes -- if it is of a source class (in
\emph{source} space), a supervised classifier is applied; otherwise,
it is in the \emph{target} space and labeled as one of the nearest
class prototypes. To allow this, it is essential to learn the compositional
spaces which separate the instances either from source and target
classes.

However, there are still two key problems remaining. First, visual
or semantic features alone may not be discriminative enough in differentiating
the source and the target classes. It is thus imperative to fully
combine the information of both visual and semantic spaces. Second,
it is notoriously difficult in tuning the model parameters for outlier
detection in judging whether an instance is from source or target
classes. Critically, a portion of instances can potentially be misclassified
regarding parameter tuning. As is illustrated in Fig.~\ref{fig:Domain-Division}
(a), we can easily find the overlapping region where the instances
may be categorized as either the source or the target classes, depending
on the model hyperparameters. This results in the misclassification
in the final prediction. Even worse, due to the aforementioned biased
problem, instances of target classes may still be inclined to be categorized
as one of source classes.

To tackle the issues, our key insight is to learn to categorize test
instances into the compositional spaces: source, target, and uncertain\footnote{As a metaphor, \emph{uncertain} here refers to the space initially taking
	as the buffer space, and finally be divided into either source or
	target spaces by our framework. } spaces. The uncertain spaces are newly introduced here to contain the
test instances that cannot be confidently classified into source or
target space, as visualized as in Fig.~\ref{fig:Domain-Division}
(a). Particularly, the source and the target space can be implicitly
learned due to the different visual/semantic distributions of source
and target classes. The recognition algorithms are applied in each
space. The uncertain space enables us to analyze the instance distribution
from a statistical perspective and we can thus categorize the class
\emph{ambiguous} instances more accurately.

Formally, we propose exploiting the distributions of source and target
classes to efficiently learn the compositional spaces from a statistical
perspective according to Fig. \ref{fig:Domain-Division}~(b). With
the extracted feature representations of images \cite{inception_v4},
our framework computes the extreme values of training instances as
confidence scores. Specifically, in term of extreme value theory \cite{scheirer2014probability},
the maximum/minimum confidence scores predicted by the classifier
of each class are drawn from extreme value distributions. Unfortunately,
we do not have prior knowledge of the underlying data distributions
of each class; thus, bootstrapping is utilized as an asymptotically
consistent method in estimating an initial boundary of source classes
and in dividing the embedding space into the source and the target
space. Nevertheless, the initial boundary estimated by bootstrapping
is too relaxed to include novel testing instances as is illustrated
in Fig.~\ref{fig:Domain-Division}. Furthermore, we introduce K-S
test \cite{massey1951kolmogorov,miller1956table,wang2003evaluating}
to validate whether the learned predictors on source classes are trustworthy,
to define the uncertain space to include instances predicted by unreliable
predictors. Intrinsically, we can take this process as the recalibration
over source space. Finally, recognition can be conducted in each learned
space. \vspace{0.05in}

\noindent \textbf{Beyond G-ZSL.} One can find that our algorithm can
be easily generalized to Open-Set Learning (OSL), which breaks\textcolor{black}{{}
	the }\textcolor{black}{\emph{closed set}}\textcolor{black}{{} assumption
	in supervised learning and recognizes the testing instances from one
	of source classes (}\textcolor{black}{\emph{i.e., source space}}\textcolor{black}{),
	or from the novel class (}\textcolor{black}{\emph{i.e., target space).}}\textcolor{black}{{}
	The novel class includes the test instances which have different distributions
	from that of the source ones. In contrast to G-ZSL, OSL only categorizes
	those instances not in source space as the novel class rather than
	a specific class.}

\vspace{0.05in}

\noindent \textbf{Contributions.} The main contribution of this paper
is to present a systematic framework in learning compositional spaces
by configuring probabilistic distributions of instances, which is
capable of addressing the G-ZSL. Towards this goal, we firstly integrate
the bootstrapping and the Kolmogorov-Smirnov test algorithms into
G-ZSL framework. In particular, we introduce a uncertain space, which
encloses the instances which cannot easily be classified into source
or target with high confidences. We extensively evaluate the importance
of compositional spaces on several G-ZSL benchmarks and achieve significant
improvement over existing G-ZSL approaches. Additionally, our framework
can be easily generalized to OSL and also achieve state-of-the-art
performances on several datasets.

\section{Related Work\label{sec:Related-Work}}

\subsection{Generalized Zero-Shot Learning}

The goal of ZSL is to recognize the instances that have never been
trained before. Typically, it requires knowledge transfer from source
to target classes where the knowledge is given in the form of semantic
attributes \cite{farhadi2009describing,fu2015transductive,lampert2009learning},
semantic word vector \cite{frome2013devise,fu2016semi,norouzi2013zero,xu2017transductive}
, or ontology \cite{rohrbach2010semantic_transfer}. Many researchers
\cite{changpinyo2016synthesized,xie2019attentive,sariyildiz2019gradient,li2019leveraging,sariyildiz2019gradient,xie2020region} recently extended ZSL to a more
general setting -- G-ZSL, where test instances can be from either
source or target classes. A thorough evaluation of G-ZSL is further
conducted by Xian \etal \cite{xian2017zero}. Their results show
that the existing ZSL algorithms do not perform well when directly
applied to G-ZSL. The predicted results are inclined to be biased
towards source classes. This is because samples of target categories
have never been reflected in the optimization of loss function, so
the model inevitably over-fits to source categories during training
time. Recent work in G-ZSL puts forward generative models to create
target instances artificially. Generative Adversarial Network (GANs)
based models \cite{chen2018zero,felix2018multi,xian2018feature} and
Variational Auto-Encoders (VAE) based models \cite{long2018pseudo,schonfeld2018generalized,verma2018generalized}
can be used for this purpose to generate examples of target classes.
However, the synthesized pseudo-samples are still not drawn from the
true sample distribution, which may interfere with both source and
target sample judgment.

Some work \cite{fu2015transductive,TransdMultilabel,transferlearningNIPS,song2018transductive}
introduced the idea of transductive learning, which utilized unlabeled
test data to help build the classification model. Particularly, these
work fine-tuned the mapping from feature space to semantic space and
update the parameters of classification models accordingly. In contrast,
though the unlabeled data are queried at the K-S test stage , note
that our framework and classification models are not updated by the
features of unlabeled test instances. Critically, the K-S Test is
a parameter-free process for statistical hypothesis testing.

\subsection{Open set Learning (OSL)}

{}{Open set Learning\footnote{In some literature, it can also be called as anomaly/novelty detection and out-of-distribution detection \cite{geng2018recent}.} focuses on the judgment whether instances are belonging to known
	classes \cite{sattar2015prediction,scheirer2014probability,scheirer2013toward,bendale2015towards}, given the testing instances. Typically, by reverting to the probability from known categories, it can judge
	whether the instances belong to unknown category by directly employing the OSL algorithms,  such as One-class Support Vector Machine (OCSVM)
	\cite{ocsvm}, Local Outlier Factor (LOF) \cite{breunig2000lof}. These algorithms utilize 
	different statistics to help the inference. Inspired by the W-SVM \cite{scheirer2014probability} in OSL, our framework firstly conducts multi-class SVM on known categories, and further utilizes extreme value theory
	to complete the task.
	Recently, there are many works on  multi-class SVM algorithms, such as the plain multi-class SVM \cite{weston1998multi}, One-against-all SVM \cite{liu2005one}, Platt's sigmoid thresholding SVM \cite{platt1999probabilistic}.
}
{}{However, the open
	set learning can only detect the instances from some unknown classes,
	rather than identifying the exact class label of instances from unknown classes. This significantly restricts its usage in real world applicationss. Unfortunately, it is very non-trivial to infer the labels of instanes in unknown classes. Semantic knowledge  should be transferred from known to unknown categories as has done in ZSL algorithms. Critically, the semantic knowledge should in principle further improve the performance of OSL in separting known from unknown data. 
	To this end, our division algorithm steps forward
	to recognize the classes of both known and unknown domains, whilst one of most important novelties comes from the newly introduced uncertain domain, with the aid of attributes to
	categorize confusing ones in feature space.}

\subsection{Bootstrapping and K-S Test}

In statistics, bootstrapping refers to random sampling with replacement
\cite{efron1994introduction}. Bootstrapping has been widely used
in machine learning, especially for bootstrap aggregating \cite{breiman1996bagging},
which is a strategy to avoid overfitting. Bootstrapping can be used
for estimating statistical properties, such as mean, variance, etc.
In our model, bootstrapping is introduced to estimate the quantile
of W-SVM confidence scores, which is more robust than a hard threshold.

K-S Test \cite{massey1951kolmogorov,miller1956table,wang2003evaluating,gretton2012kernel}
is one of the most well-known test to examine whether two samples
are drawn from the same distribution. It validates the distance between
the Cumulative Distribution Functions (CDF), which can be recognized
as a metric distance. In transfer learning, such a distance is essential
for domain adaptation. Long \etal \cite{long2015learning} firstly
introduced two-sample test in transfer learning, which aims to shorten
the distance between source and target space. In our setting, we would
like to identify the distance and classify them respectively, which
is different from previous work.

\section{Learning Compositional Spaces for G-ZSL \label{sec:Domain-Division-Algorithm}}

\noindent \textbf{Problem setup.} In our learning task, we have a
training dataset\emph{, i.e.}, source classes, with $n_{s}$ instances,
$\mathcal{D}_{s}=\{\mathbf{x}_{i},\mathbf{y}_{i},l_{i}\}_{i=1}^{n_{s}}$:
$\mathbf{x}_{i}\in\mathbb{R}^{n}$ is the feature of $i_{th}$ instance
with the class label $l_{i}\in\mathcal{C}_{s}$, where $\mathcal{C}_{s}$
is the source class set; $n_{s}^{c}$ is the number of instances in
source class $c$. Analogous to standard ZSL setting, we introduce
target label classes $\mathcal{C}_{t}$ with $\mathcal{C}_{s}\bigcap\mathcal{C}_{t}=\emptyset$
and the full class label set $\mathcal{C}=\mathcal{C}_{s}\cup\mathcal{C}_{t}$.
$\mathbf{y}_{i}$ is the semantic vector of instance $\mathbf{x}_{i}$.
According to Lampert \etal \cite{lampert2014attribute}, we assume
a class-level semantic vector profile existed: $\mathbf{y}_{c}$ is
denoted as the semantic prototype for all the instances in class $c$.
Given one test instance $\mathbf{x}_{i}$, our goal is to predict
its class label $c_{i}$. In G-ZSL tasks, we target at learning to
predict $c_{i}\in\left\{ \mathcal{C}_{s},\mathcal{C}_{t}\right\} $.
The semantic prototype is predefined for each class in $\mathcal{C}_{t}$.
Additionally, our framework can also address the OSL task by predicting
$c_{i}\in\left\{ \mathcal{C}_{s},novel\,class\right\} $, where the
\emph{novel class} is an umbrella term referring to any class not
in $\mathcal{C}_{s}$. 

\subsection{Extreme Values as Confidence Scores }

One can conduct the G-ZSL by directly learning compositional spaces
of source and target classes. Such an idea has been well explored
in the CMT \cite{socher2013zero}, which employed Local Outlier Factor
(LOF) to measure the degree a data point is outlying in target space.
Thus, CMT relies on the density of clusters of source classes. In
contrast, our framework of learning compositional space is developed
upon the extreme value theory \cite{scheirer2014probability}.

Given a test instance $\mathbf{x}_{i}$, the supervised predictor
can compute the confidence score $z_{_{i}}^{c}=f_{c}\left(\mathbf{x}_{i}\right)$\footnote{In our setting, we use W-SVM to get the score, whose input is the neural network pre-trained feature. Note that the score is the value before softmax prediction.},
which indicates the certainty of $\mathbf{x}_{i}$ belonging to the
class $c$. In our experiments, we follow the setting of W-SVM, which utilize the output of Weibull-based SVM as the confidence predictor.
Therefore, we define two events,

\[
\begin{cases}
	E_{1}: & \mathbf{x}_{i}\:\mathrm{belonging\:to\:class\:}c;\\
	E_{2}: & \mathbf{x}_{i}\ \mathrm{belonging\:to\:the\:other\:classes.}
\end{cases}
\]

In term of extreme value theory \cite{scheirer2014probability}, the
maximum/minimum confidence scores predicted by the classifier of each
class can be taken as one of extreme value distributions {}{(\emph{i.e.},
	Weibull distribution $G(z;\lambda_c,\nu_c,\kappa_c)$  and reverse Weibull distributions $rG(z;\lambda_c',\nu_c',\kappa_c')$ respectively). Note that the parameters of weibull/reverse-weibull density
	$\lambda_c,\nu_c,\kappa_c,\lambda_c',\nu_c',\kappa_c'$ can be obtained by MLE, which was introduced by W-SVM.}
We can thus estimate
the upper/lower extremes of instance $\mathbf{x}_{i}$ for event $E_{1}$/$E_{2}$
individually. {}{Accordingly, given $\mathbf{x}_{i}$ and $z_{_{i}}^{c}$,
	we can estimate the probability of these two events as $P_G\left(E_{1}\right)=1-\exp\left(-\left(\frac{z_{i}^{c}-\nu_{c}}{\lambda_{c}}\right)^{\kappa_{c}}\right)$
	and $P_{rG}\left(\lnot E_{2}\right)=\exp\left(-\left(\frac{}{}\frac{z_{i}^{c}-\nu'_{c}}{\lambda'_{c}}\right)^{\kappa'_{c}}\right)$, note that $G$ and $rG$ refer to the weibull distribution and reverse weibull distribution, $P_G$ and $P_{rG}$ represent their probability.}

We introduce the calibrated extreme values $m_{c}\left(\mathbf{x}_{i}\right)$
\cite{scheirer2014probability} as the \emph{confidence scores} in
measuring the confidence that $x_{i}$ belonging to class $c$ as,

\begin{equation}
	m_{c}\left(\mathbf{x}_{i}\right)=P_{rG}\left(\lnot E_{2}\mid f_{c}\left(\mathbf{x}_{i}\right)\right)\cdot P_{G}\left(E_{1}\mid f_{c}\left(\mathbf{x}_{i}\right)\right)\label{eq:multiply}
\end{equation}

To determine if one testing instance $\mathbf{x}_{i}$ belongs to class
$c$, Weibull-calibrated SVM (W-SVM) \cite{scheirer2014probability}
introduced a threshold $\delta_{c}$ as

\begin{equation}
	c_{i}=\begin{cases}
		c & m_{c}\left(\mathbf{x}_{i}\right)>\delta_{c}\\
		\mathrm{\lnot c} & \mathrm{otherwise}
	\end{cases}\label{eq:threshold}
\end{equation}

\noindent where $\delta_{c}$ is a fixed value \cite{scheirer2014probability}
in determining whether the instance $i$ belongs to the class $c$.
The instance $\mathbf{x}_{i}$ rejected by all source classes in Eq
(\ref{eq:threshold}) should be labeled as the target space. Generalizing
to $\mathcal{C}_{s}$ class is straightforward by training multiple
prediction functions $\left\{ f_{c}\left(\mathbf{x}\right)\right\} $,
$c=1,\cdots,\left|\mathcal{C}_{s}\right|$.

\vspace{0.05in}

\noindent \textbf{Limitations.} We argue that there are several key
limitations indirectly utilizing Eq (\ref{eq:multiply}) and Eq (\ref{eq:threshold})
of learning compositional spaces: First, it is undesirable to have
a fixed threshold $\delta_{c}$ which is empirically pre-defined for
any data distributions in the source space, since the score is not
an invariant. Essentially, the instances may derive from many different
source classes. Such a fixed threshold cannot account the versatile
data distribution in practice. For example, the instances in our uncertain
space (Fig. \ref{fig:Domain-Division}) may be predicted by a wrong
source/target space label; these wrongly labeled instances will never
be correctly categorized. Furthermore, Eq (\ref{eq:multiply}) directly
multiply two terms, which presumes a potential hypothesis that no
correlation exists between $E_{1}$ and $\lnot E_{2}$, which is not
the case in reality. 

\subsection{Model Selection by Bootstrapping\label{subsec:Determining-the-Initial}}

Rather than using a fixed $\delta_{c}$ , it can be taken as a model
selection task in estimating the $\delta_{c}$ for Eq (\ref{eq:multiply})
and Eq (\ref{eq:threshold}). Therefore, we tackle this task with
the bootstrapping approach \cite{bootstrap}. Different from the \emph{bootstrapping}
(\emph{i.e.}, self-training) in computer vision \cite{ShrivastavaECCV12},
bootstrapping is a statistical strategy in estimating the statistics
of sampling distributions. Particularly, it estimates the standard
errors and the confidence intervals of parameters of the underlying
distributions. Such a procedure essentially enables model selection
to determine the parameters.

Its procedures are closely related to the other methods such as cross-validation
and jackknife sampling. The whole algorithm is shown in Alg. \ref{alg:Determining-the-initial}.
To facilitate the discussion, we denote the training set of class
$c$ as $\left\{ \mathbf{x}_{tr}^{c}\right\} $; the testing set whose
instances are mostly confidently predicted as class $c$, as $\left\{ \mathbf{x}_{te}^{c}\right\} $.
Thus, the corresponding confidence score set on training and testing
data are $\left\{ {z}_{tr}^{c}\right\} =m_{c}\left(\left\{ \mathbf{x}_{tr}^{c}\right\} \right)$
and $\left\{ {z}_{te}^{c}\right\} =m_{c}\left(\left\{ \mathbf{x}_{te}^{c}\right\} \right)$
respectively. We can sample the $\left\{ {z}_{tr}^{c}\right\} $ (with
replacement) to obtain the quantile of confidence scores for each
class, then the threshold can be defined directly.

\begin{algorithm}
	\begin{description}
		\item [{{\small{}Input:}}] {\small{}{}Confidence score set on training
			data $\left\{ {z}_{tr}^{c}\right\} $}{\small\par}

		\item [{{\small{}Output:}}] {\small{}{}Threshold $\delta_{c}$: }{\small\par}

	\end{description}
	\begin{enumerate}
		\item {\small{}{}We sample from $\left\{ {z}_{tr}^{c}\right\} $ for $n$
			times (with replacement), producing a sampling set $\left\{ \widetilde{{z}}_{tr\left(k\right)}^{c}\right\} _{k=1}^{n}$,
			where $\widetilde{{z}}_{tr\left(k\right)}^{c}$ indicates the $k_{th}$
			sampled instance; }{\small\par}

		\item {\small{}{}We also choose the significance level $\alpha$, and generate
			the $\alpha$ quantile $\widetilde{{z}}_{tr}^{c\star}$ from $\left\{ \widetilde{{z}}_{tr\left(k\right)}^{c}\right\} _{k=1}^{n}$.
			Particularly, we sort $\widetilde{{z}}_{tr\left(k\right)}^{c}$ with
			an ascending order and extract $\left(\max\left(\mathrm{Round}\left[\alpha n\right],1\right)\right)_{th}$
			value as $\widetilde{{z}}_{tr}^{c\star}$. We repeat it for $n$ times
			over $\left\{ {z}_{tr}^{c}\right\} $ to get $\left\{ \widetilde{{z}}_{tr\left(k\right)}^{c\star}\right\} _{k=1}^{n}$. }{\small\par}

		\item {\small{}{}The threshold of Eq (\ref{eq:threshold}) can thus be
			computed as the mean of these values, }\emph{\small{}{}i.e., }{\small{}{}$\delta_{c}=\frac{1}{n}\Sigma_{k=1}^{n}\widetilde{{z}}_{tr\left(k\right)}^{c\star}$.\caption{\label{alg:Determining-the-initial}Determining the initial threshold}
		}{\small\par}

	\end{enumerate}
\end{algorithm}

Till now, we had a sketch of our algorithm. Specifically, the training
instances of source classes are utilized to get the $m_{c}\left(\cdot\right)$,
$c\in\mathcal{C}_{s}$; For any given testing instance $\mathbf{x}_{i}$,
we compute its confidence score $m_{c}\left(\mathbf{x}_{i}\right)$,
$c\in\mathcal{C}_{s}$. To determine whether an instance $\mathbf{x}_{i}$
is in source or target classes, we calculate the statistic $m_{c}\left(\mathbf{x}_{i}\right)$
in Eq (\ref{eq:multiply}) with the threshold $\delta_{c}$ estimated
by the bootstrapping algorithm in Alg. \ref{alg:Determining-the-initial}.
The instances computed in the source and target space will be categorized
by supervised, or zero-shot classifiers respectively. Once the $\delta_{c}$
is estimated, we can have the boundary between source and target spaces.

Note that the whole framework relies on the classifier $m_{c}\left(\cdot\right)$,
$c\in\mathcal{C}_{s}$ which is supposed to be robust and well-trained.
Unfortunately empirically, we cannot always train good classifiers
for all classes, \eg, the class with insufficient training samples.
It is also nontrivial to tune the hyper-parameters of these predictors,
especially, the deep network based predictors.

\subsection{Learning Uncertain Space by K-S Test\label{subsec:Fine-tuning-the-Threshold}}

The bootstrapping in Alg. \ref{alg:Determining-the-initial} only
give a good approximation of the distributions of empirical quantiles
in practice \cite{bootstrap_weakness}. In G-ZSL task, we observe
that the estimated $\delta_{c}$ may be consistently too relaxed to
determine the boundary of the source space. As a result, we have to 
establish an \emph{uncertain} part for better prediction.

The key idea of updating by bootstrapping is to validate whether the
learned classifier $m_{c}\left(\cdot\right)$, $c\in\mathcal{C}_{s}$
is trustworthy. Generally, we assume the instances of class $c$ independent
and identically distributed. Given sufficient training samples, an
ideal classifier $m_{c}\left(\cdot\right)$ should produce similar
confidence score distributions of training and testing instances\footnote{A good $m_{c}\left(\cdot\right)$ and $\left\{ \mathbf{x}_{tr}^{c}\right\} $
	should be conditionally independent given the distribution of each
	class $c$, which can be controlled by regularization term} of class $c$. The Kolmogorov-Smirnov (K-S) test is an efficient,
straightforward, and qualified choice method for comparing distributions
\cite{massey1951kolmogorov,miller1956table,wang2003evaluating}. Remarkably,
K-S test is a distribution-free test and the statistics of K-S test
is effortless to compute. We define the null and alternative hypothesis
as 
\begin{equation}
	\!\!\!\!\!\!\begin{split} & H_{0}\mathrm{:\mathit{\{{z}_{tr}^{c}\}}\ and\ \mathit{\{{z}_{te}^{c}\}}\ are\ from\ the\ same\ distribution}.\\
		& H_{1}\mathrm{:\mathit{\{{z}_{tr}^{c}\}}\ and\ \mathit{\{{z}_{te}^{c}\}}\ are\ from\ different\ distributions}.
	\end{split}
	\label{eq:h1}
\end{equation}

We define $K^{c}=\underset{{z}}{\mathrm{sup}}\left\Vert F_{tr}^{c}\left({z}\right)-F_{te}^{c}\left({z}\right)\right\Vert $
as the distance measure, where $F_{tr}^{c}=\mathrm{ecdf}(\{{z}_{tr}^{c}\})$
and $F_{te}^{c}=\mathrm{ecdf}(\{{z}_{te}^{c}\})$; $\mathrm{ecdf}\left(\cdot\right)$
is the empirical distribution function. The null hypothesis would
be rejected at the significant level $\alpha$ when,

\begin{equation}
	K^{c}(\alpha)>\sqrt{-\frac{\left|\{{z}_{tr}^{c}\}\right|+\left|\{{z}_{te}^{c}\}\right|}{2\left|\{{z}_{tr}^{c}\}\right|\cdot\left|\{{z}_{te}^{c}\}\right|}\log\left(\frac{\alpha}{2}\right)}.\label{eq:alpha}
\end{equation}

\noindent When $H_{0}$ is accepted, it indicates that the $m_{c}\left(\cdot\right)$
is trustworthy, and the confidence scores of training and testing
instances in class $c$ come from the same distribution. We are certain
that a large portion of testing instances $\left\{ {z}_{te}^{c}\right\} =m_{c}\left(\left\{ \mathbf{x}_{te}^{c}\right\} \right)$
should be in the class $c$. On the other hand, when $H_{0}$ is rejected,
we are not sure whether $m_{c}\left(\cdot\right)$ is well learned;
the class labels of these testing instances are uncertain. To this
end, we introduce a new compositional space -- uncertain space to include
these instances.

\vspace{0.05in}

\noindent \textbf{Uncertain Space.} Labels of instances in the uncertain
space should be labeled as the most likely source class or one of
target classes. Specifically, we can compute the $\left\{ z^{c}=m_{c}\left(\mathbf{x}\right)\right\} _{c=1}^{\left|\mathcal{C}_{s}\right|}$
over all $\mathcal{C}_{s}$ classes; and we can obtain, 
\begin{equation}
	\begin{cases}
		c^{\star}=\mathrm{argmax}_{c\in\mathcal{C}_{s}}\:\left\{ z^{c}\right\} \\
		z^{\star}=\mathrm{max}_{c\in\mathcal{C}_{s}}\:\left\{ z^{c}\right\} 
	\end{cases}\label{eq:optimised_z}
\end{equation}

The mapping function $g\left(\cdot\right)$ is learned on the source
space from features $\mathbf{x}_{i}$ to its corresponding semantic
vector $\mathbf{y}_{i}$. Given one testing instance $\mathbf{x}_{i}$:
if $z_{i}^{\star}$ is very high, we can confidently predict $\mathbf{x}_{i}$
belonging to one of source classes; otherwise, the label of $\mathbf{x}_{i}$
is either in the uncertain or target space. We thus have,

\begin{equation}
	c_{i}^{\star}=\underset{c\in\mathcal{C}_{t}\cup\{c^{\star}\}}{\mathrm{argmin}}\left\Vert g\left(\mathbf{x}_{i}\right)-\mathbf{y}_{c}\right\Vert \label{eq:uncertain}
\end{equation}

\noindent where $\mathbf{y}_{c}$ is the semantic prototype of class
$c$; $c^{\star}$ is the most likely source class to which $\mathbf{x}_{i}$
belongs to.

{}{\noindent \textbf{Sample Selections.} In general, we have three spaces so far: \emph{source},  \emph{target}, and  \emph{uncertain}. We firstly use the superviser predictor $f_c$ to obtain the confidence score $z_i^c$, then use initial threshold determined by Alg.\ref{alg:Determining-the-initial} to split rough \emph{source} and \emph{target} space. Finally, we use K-S test to bring \emph{uncertain} space to fine-tune the embedding space.}

\section{Recognition in Compositional Spaces}

With the learned compositional spaces, we can make predictions in
source, target, and uncertain space. Formally, we make the prediction
as,

\begin{equation}
	c_{i}^{\star}=\begin{cases}
		\underset{c\in\mathcal{C}_{s}}{\mathrm{argmax}}\ m_{c}\left(\mathbf{x}_{i}\right) & \mathrm{Source\ space}\\
		\underset{c\in\mathcal{C}_{t}}{\mathrm{argmin}}\left\Vert g\left(\mathbf{x}_{i}\right)-\mathbf{y}_{c}\right\Vert  & \mathrm{Target\ space}\\
		\underset{c\in\mathcal{C}_{t}\cup\{c^{\star}\}}{\mathrm{argmin}}\left\Vert g\left(\mathbf{x}_{i}\right)-\mathbf{y}_{c}\right\Vert  & \mathrm{Uncertain\ space}
	\end{cases}\label{eq:all_z}
\end{equation}

Thus, we use $m_{c}\left(\mathbf{x}_{i}\right)$ to determine which
space $\mathbf{x}_{i}$ belongs to. If $\mathbf{x}_{i}$ is from the
source space, the label can be directly got by $m_{c}\left(\mathbf{x}_{i}\right)$;
if $\mathbf{x}_{i}$ is from the target space, we should draw support
from $g\left(\mathbf{x}_{i}\right)$ and $\mathbf{y}_{c}$ to predict
the label. Moreover, the search space of target space differs, due
to the dissimilar likelihood for $\mathbf{x}_{i}$ from source classes.
Additionally, a state-of-the-art supervised classifier $f\left(\cdot\right)$
and a zero-shot learner $g\left(\cdot\right)$ are orthogonal and
potential useful here, since our work is a general framework and we
do not define the specific forms of these classifiers.

\noindent \textbf{Target versus Uncertain Spaces.} We highlight the
differences between these two spaces. Particularly, by using the learned
embedding $g\left(\cdot\right)$, the class labels of instances can
be inferred as, 
\begin{equation}
	\begin{cases}
		c\in\mathcal{C}_{t} & \mathrm{Target\ space}\\
		c\in\mathcal{C}_{t}\cup\{c^{\star}\} & \mathrm{Uncertain\ space}
	\end{cases}\label{eq:c}
\end{equation}

\noindent where $c^{\star}$ is the most likely source class for $\mathbf{x}_{i}$
which is computed by the supervised classifier. Therefore, the search
space of our framework is $\left|\mathcal{C}_{t}\right|$ (target
space) or $\left|\mathcal{C}_{t}\right|+1$ (Uncertain space), rather
than $\left|\mathcal{C}_{t}\right|+\left|\mathcal{C}_{s}\right|$
in \cite{wild_0shot}.

\subsection{Recognition in Source Space}

As the $m_{c}\left(\mathbf{x}_{i}\right)$ is available for any $c\in\mathcal{C}_{s}$,
we can directly find the most likely class for $\mathbf{x}_{i}$ as
our prediction. In particular, we employ the $\mathrm{argmax}_{\mathrm{c\in\mathcal{C}_{s}}}\ m_{c}\left(\mathbf{x}_{i}\right)$
as prediction of our model in the source space. SVM is used here.

\subsection{Recognition in Target and Uncertain Spaces}

The predictions in target/uncertain space entail a good embedding from
feature space to semantic space, \emph{i.e.}, $g\left(\cdot\right)$.
In general, $g\left(\cdot\right)$ should be flexible to use any ZSL
algorithm. Particularly, we adopt the linear and non-linear embedding
ZSL algorithms in our framework.

\vspace{0.05in}

\noindent \textbf{Linear Embedding: } The linear model is utilized
in learning the embedding. Impressively, such a plain model can achieve
remarkable results compared against the other G-ZSL algorithms, as
shown in the experimental section. Particularly, we propose the directly
mapping model (D-M) which just employs the feature prototypes of each
class and a linear predictor in predicting the attribute/word vector
$g\left(\mathbf{x}\right)=\mathbf{w}^{T}\cdot\mathbf{x}$. The feature
prototype embedding is computed as,

\begin{equation}
	\mathbf{w}=\underset{\mathbf{w}}{\mathrm{argmin}}\sum_{c\in\mathcal{C}_{s}}\left\Vert g(\widetilde{\mathbf{x}}_{c})-\mathbf{y}_{c}\right\Vert +\lambda\left\Vert \mathbf{w}\right\Vert _{2}
\end{equation}

\noindent where $\widetilde{\mathbf{x}}_{c}$ is feature prototype
of class $c$ computed by averaged the instance features of source
class $c$; $\mathbf{y}_{c}$ is the semantic prototype of class $c$,
with the computed embedding weight $\mathbf{w}$.

\vspace{0.05in}

\noindent \textbf{Non-linear Embedding:} To further show the efficacy
of our framework, we also consider the non-linear embedding model.
The Adversarial Generative Model (A-G) is investigated here, since
the generative models can better learn the feature embedding. Particularly,
we implement the \emph{f-CLSWGAN} \cite{xian2018feature} as the algorithm
for target domain.

\vspace{0.05in}

\noindent \textbf{Beyond G-ZSL. }Our framework can be generalized
to Open-Set Learning, by predicting the labels as,

\begin{equation}
	\hat{c}^{\star}=\begin{cases}
		\hat{c} & \hat{c}\in\mathcal{C}_{s}\\
		novel\ class & \mathrm{otherwise}
	\end{cases}\label{eq:all_z-1}
\end{equation}

\noindent where $\hat{c}$ is the predicted class label in source
space, and we denote the instances not belonging to any source class
as the $novel\ class$.

\section{Experiments}

\subsection{Datasets and settings\label{subsec:Datasets-and-settings}}

\noindent 
\begin{table*}
	\begin{centering}
		{\small{}{}\caption{G-ZSL Results on AwA, CUB and aPY. ($^{\star}$ Our implement, $Acc$ refers to accuracy (\%), $H$ is the harmonic mean)\label{tab:GZSL}}
		}{\small\par}
		\par\end{centering}
	\begin{centering}
		
		\par\end{centering}
	\centering{}%
	\begin{tabular}{c|c|ccc|ccc|ccc}
		\hline 
		\multirow{1}{*}{{\small{}{}Type}} & \multirow{1}{*}{{\small{}{}Method}} & \multicolumn{3}{c|}{{\small{}{}AwA}} & \multicolumn{3}{c|}{{\small{}{}CUB }} & \multicolumn{3}{c}{{\small{}{}aPY}}\tabularnewline
		&  & {\small{}{}$Acc_{\mathbb{U}\rightarrow\mathbb{T}}$ } & {\small{}{}$Acc_{\mathbb{S}\rightarrow\mathbb{T}}$ } & {\small{}{}$H$ } & {\small{}{}$Acc_{\mathbb{U}\rightarrow\mathbb{T}}$ } & {\small{}{}$Acc_{\mathbb{S}\rightarrow\mathbb{T}}$ } & {\small{}{}$H$  } & {\small{}{}$Acc_{\mathbb{U}\rightarrow\mathbb{T}}$ } & {\small{}{}$Acc_{\mathbb{S}\rightarrow\mathbb{T}}$ } & {\small{}{}$H$}\tabularnewline
		\hline 
		\multirow{12}{*}{{\small{}ZSL Models}} & {\small{}{}Chance } & {\small{}{}2.0 } & {\small{}{}2.0 } & {\small{}{}- } & {\small{}{}0.5 } & {\small{}{}0.5 } & {\small{}{}- } & {\small{}{}3.1 } & {\small{}{}3.1 } & {\small{}{}-}\tabularnewline
		& {\small{}{}DAP } & {\small{}{}0.0 } & {\small{}{}{88.7} } & {\small{}{}{0.0} } & {\small{}{}1.7 } & {\small{}{}67.9 } & {\small{}{}{3.3} } & {\small{}{}4.8 } & {\small{}{}78.3 } & {\small{}{}9.0}\tabularnewline
		& {\small{}{}ConSE } & {\small{}{}0.4 } & {\small{}{}{88.6} } & {\small{}{}{0.8} } & {\small{}{}{1.6} } & \textbf{\small{}{}72.2}{\small{} } & {\small{}{}{3.1} } & {\small{}{}0.0 } & \textbf{\small{}{}91.2}{\small{} } & {\small{}{}0.0}\tabularnewline
		& {\small{}{}CMT } & {\small{}{}8.4 } & {\small{}{}{86.9} } & {\small{}{}{15.3} } & {\small{}{}4.7 } & {\small{}{}{60.1} } & {\small{}{}{8.7} } & {\small{}{}10.9 } & {\small{}{}74.2 } & {\small{}{}19.0}\tabularnewline
		& {\small{}{}SSE } & {\small{}{}7.0 } & {\small{}{}{80.6} } & {\small{}{}{12.9} } & {\small{}{}{8.5} } & {\small{}{}{46.9} } & {\small{}{}{14.4} } & {\small{}{}0.2 } & {\small{}{}78.9 } & {\small{}{}0.4}\tabularnewline
		& {\small{}{}Latem } & {\small{}{}7.3 } & {\small{}{}{71.7} } & {\small{}{}{13.3} } & {\small{}{}{15.2} } & {\small{}{}{57.3} } & {\small{}{}{24.0} } & {\small{}{}0.1 } & {\small{}{}73.0 } & {\small{}{}0.2}\tabularnewline
		& {\small{}{}ALE } & {\small{}{}16.8 } & {\small{}{}{76.1} } & {\small{}{}{27.5} } & {\small{}{}{23.7} } & {\small{}{}{62.8} } & {\small{}{}{34.4} } & {\small{}{}4.6 } & {\small{}{}73.7 } & {\small{}{}8.7}\tabularnewline
		& {\small{}{}DeViSE } & {\small{}{}13.4 } & {\small{}{}{68.7} } & {\small{}{}{22.4} } & {\small{}{}{23.8} } & {\small{}{}{53.0} } & {\small{}{}{32.8} } & {\small{}{}4.9 } & {\small{}{}76.9 } & {\small{}{}9.2}\tabularnewline
		& {\small{}{}SJE } & {\small{}{}11.3 } & {\small{}{}{74.6} } & {\small{}{}{19.6} } & {\small{}{}{23.5} } & {\small{}{}{59.2} } & {\small{}{}{33.6} } & {\small{}{}3.7 } & {\small{}{}55.7 } & {\small{}{}6.9}\tabularnewline
		& {\small{}{}ESZSL } & {\small{}{}6.6 } & {\small{}{}{75.6} } & {\small{}{}{12.1} } & {\small{}{}{12.6}  } & {\small{}{}{63.8} } & {\small{}{}{21.0} } & {\small{}{}2.4 } & {\small{}{}70.1 } & {\small{}{}4.6}\tabularnewline
		& {\small{}{}SYNC } & {\small{}{}8.9 } & {\small{}{}87.3 } & {\small{}{}16.2 } & {\small{}{}11.5 } & {\small{}{}{70.9} } & {\small{}{}{19.8} } & {\small{}{}7.4 } & {\small{}{}66.3 } & {\small{}{}13.3}\tabularnewline
		& {\small{}{}SAE } & {\small{}{}1.1 } & {\small{}{}82.2 } & {\small{}{}2.2 } & {\small{}{}7.8 } & {\small{}{}{54.0} } & {\small{}{}13.6 } & {\small{}{}0.4 } & {\small{}{}80.9 } & {\small{}{}0.9 }\tabularnewline
		\hline 
		\multirow{8}{*}{{\small{}G-ZSL Models }} & \textcolor{black}{\small{}{}SE-GZSL}{\small{} } & \textcolor{black}{\small{}{}56.3}{\small{} } & \textcolor{black}{\small{}{}67.8}{\small{} } & \textcolor{black}{\small{}{}61.5}{\small{} } & \textcolor{black}{\small{}{}41.5}{\small{} } & \textcolor{black}{\small{}{}53.3}{\small{} } & \textcolor{black}{\small{}{}46.7}{\small{} } & \textcolor{black}{\small{}{}-}{\small{} } & \textcolor{black}{\small{}{}-}{\small{} } & \textcolor{black}{\small{}{}-}\tabularnewline
		& {\small{}{}CADA-VAE } & {\small{}{}57.3 } & {\small{}{}72.8 } & {\small{}{}64.1 } & {\small{}{}51.6 } & {\small{}{}53.5 } & {\small{}{}52.4 } & {\small{}{}- } & {\small{}{}- } & {\small{}{}-}\tabularnewline
		& \textcolor{black}{\small{}{}PTMCA}{\small{} } & \textcolor{black}{\small{}{}22.4}{\small{} } & \textcolor{black}{\small{}{}80.6}{\small{} } & \textcolor{black}{\small{}{}35.1}{\small{} } & \textcolor{black}{\small{}{}23.0}{\small{} } & \textcolor{black}{\small{}{}51.6}{\small{} } & \textcolor{black}{\small{}{}31.8}{\small{} } & \textcolor{black}{\small{}{}15.4}{\small{} } & \textcolor{black}{\small{}{}71.3}{\small{} } & \textcolor{black}{\small{}{}25.4}\tabularnewline
		& {\small{}{}SP-AEN } & {\small{}{}23.3 } & {\small{}{}90.9 } & {\small{}{}37.1 } & {\small{}{}34.7 } & {\small{}{}70.6 } & {\small{}{}46.6 } & {\small{}{}13.7 } & {\small{}{}63.4 } & {\small{}{}22.6}\tabularnewline
		& \textcolor{black}{\small{}{}cycle-CLSWGAN}{\small{} } & \textcolor{black}{\small{}{}56.9}{\small{} } & \textcolor{black}{\small{}{}64.0}{\small{} } & \textcolor{black}{\small{}{}60.2}{\small{} } & \textcolor{black}{\small{}{}45.7}{\small{} } & \textcolor{black}{\small{}{}61.0}{\small{} } & \textcolor{black}{\small{}{}52.3}{\small{} } & \textcolor{black}{\small{}{}-}{\small{} } & \textcolor{black}{\small{}{}-}{\small{} } & \textcolor{black}{\small{}{}-}\tabularnewline
		& \textcolor{black}{\small{}{}f-CLSWGAN}{\small{} } & \textcolor{black}{\small{}{}57.9}{\small{} } & \textcolor{black}{\small{}{}61.4}{\small{} } & \textcolor{black}{\small{}{}59.6}{\small{} } & \textcolor{black}{\small{}{}43.7}{\small{} } & \textcolor{black}{\small{}{}57.7}{\small{} } & \textcolor{black}{\small{}{}49.7}{\small{} } & \textcolor{black}{\small{}{}-}{\small{} } & \textcolor{black}{\small{}{}-}{\small{} } & \textcolor{black}{\small{}{}-}\tabularnewline
		& \textcolor{black}{\small{}{}f-CLSWGAN}{\small{}{}$^{\star}$ } & {\small{}{}57.8 } & {\small{}{}72.4 } & {\small{}{}64.2 } & {\small{}{}43.4 } & {\small{}{}58.3 } & {\small{}{}49.8 } & {\small{}{}16.8 } & {\small{}{}45.7 } & {\small{}{}24.6}\tabularnewline
		& {\small{}{}CDL } & {\small{}{}28.1 } & {\small{}{}73.5 } & {\small{}{}40.6 } & {\small{}{}23.5 } & {\small{}{}55.2 } & {\small{}{}32.9 } & {\small{}{}19.8 } & {\small{}{}48.6 } & {\small{}{}28.1}\tabularnewline
		\hline 
		\multirow{2}{*}{{\small{}Ours}} & {\small{}{}SVM/D-M } & {\small{}{}53.6 } & {\small{}{}90.4 } & {\small{}{}67.3 } & {\small{}{}37.2 } & {\small{}{}45.2 } & {\small{}{}40.8 } & \textbf{\small{}{}44.0}{\small{} } & {\small{}{}89.2 } & \textbf{\small{}{}58.9}\tabularnewline
		& {\small{}{}SVM/A-G } & \textbf{\small{}{}66.0}{\small{} } & \textbf{\small{}{}91.2}{\small{} } & \textbf{\small{}{}76.6}{\small{} } & \textbf{\small{}{}53.1}{\small{} } & {\small{}{}59.4 } & \textbf{\small{}{}56.1}{\small{} } & {\small{}{}22.4 } & {\small{}{}81.3 } & {\small{}{}35.1}\tabularnewline
		\hline 
	\end{tabular}
\end{table*}

\noindent \textbf{Datasets.} \emph{Animal with Attribute (AwA)} Dataset
\cite{lampert2014attribute} has 50 classes and 30,475 images in total,
with 85 class-level attributes annotated. We use 40 source training
classes (including 13 classes as validation); the rest are for testing.
\emph{CUB} Dataset \cite{WahCUB_200_2011} includes 200 classes and
11,788 fine-grain images with 312 class-level attributes annotated.
The training set has 150 classes (including 50 classes as validation).
(3) \emph{aPY} Dataset \cite{farhadi2009describing} has 15,339 images
in 32 classes with 64 class-level annotated attributes. We use 20
classes for training (including 5 validation classes). {}{For the AwA,
	CUB and aPY, we use ResNet-101 features and the class split contributed
	by Xian \etal \cite{xian2017zero}.} (4) ImageNet 2012/2010 dataset
is proposed by Fu \etal \cite{fu2016semi}. {}{As the large-scale dataset,
	we use the VGG-19 feature and split as Fu \etal \cite{fu2016semi}}: 1000 training classes
with full training instances in ILSVRC 2012; and 360 testing classes
in ILSVRC 2010, non-overlapped with ILSVRC 2012 classes. {}{Notably, the 
	attribute of ILSVRC dataset is the word2vec vectors provided by \cite{fu2016semi}.}

\noindent 
\begin{table*}[h!]
	\caption{\label{tab:G-ZSL-Imagenet}G-ZSL on the large-scale dataset -- ImageNet
		2012/2010. ($Acc$ refers to accuracy (\%), $H$ is the harmonic mean)}
	
	\centering{}%
	\begin{tabular}{l|cccccc|c}
		\hline 
		& {\small{}SS-Voc } & {\small{}SAE } & {\small{}ESZSL } & {\small{}DeViSE } & {\small{}ConSE } & {\small{}Chance } & {\small{}Ours (SVM/D-M)}\tabularnewline
		\hline 
		{\small{}{}$Acc_{\mathbb{U}\rightarrow\mathbb{T}}$ } & {\small{}2.3 } & {\small{}0.2 } & {\small{}0.5 } & {\small{}0.4 } & {\small{}0.0 } & {\small{}$<$0.1 } & \textbf{\small{}5.7}{\small{} }\tabularnewline
		{\small{}{}$Acc_{\mathbb{S}\rightarrow\mathbb{T}}$ } & {\small{}33.5 } & {\small{}32.8 } & {\small{}38.1 } & {\small{}24.7 } & \textbf{\small{}56.2}{\small{} } & {\small{}$<$0.1 } & {\small{}54.1 }\tabularnewline
		{\small{}$H$ } & {\small{}4.3 } & {\small{}0.5 } & {\small{}0.9 } & {\small{}0.8 } & {\small{}0.0 } & {\small{}- } & \textbf{\small{}10.3}{\small{} }\tabularnewline
		\hline 
	\end{tabular}
\end{table*}

\vspace{0.05in}

\noindent \textbf{Experimental settings.} Our model is validated in
standard G-ZSL settings as \cite{xian2017zero}. G-ZSL gives the class label of testing instances either
from source or target classes. We set the significance level $\alpha=0.05$
to tolerate $5\%$ Type-I error. By default, we use SVM with RBF kernel
with parameter cross-validated, unless otherwise specified. The code
will be available once acceptance.

\subsection{Results of Generalized Zero-Shot Learning}

\noindent \textbf{Settings}: We first compare the experiments on G-ZSL
by using the settings by Xian \etal \cite{xian2017zero}. The results
are summarized in Tab.~\ref{tab:GZSL}. {}{In particular, we further
	compare the separate settings; and top-1 accuracy in ($\%$) is reported
	here}: (1) $Acc_{\mathbb{S}\rightarrow\mathbb{T}}$: Test instances
from source classes, the prediction candidates include both source
and target classes; (2) $Acc_{\mathbb{U}\rightarrow\mathbb{T}}$:
Test instances from target classes, the prediction candidates include
both source and target classes. (3) {}{We employ the harmonic mean as
	the main evaluation metric to further combine the results of both
	$\mathbb{S}\rightarrow\mathbb{T}$ and $\mathbb{U}\rightarrow\mathbb{T}$,
	as $H=2\cdot Acc_{\mathbb{U}\rightarrow\mathbb{T}}\cdot Acc_{\mathbb{S}\rightarrow\mathbb{T}}/\left(Acc_{\mathbb{U}\rightarrow\mathbb{T}}+Acc_{\mathbb{S}\rightarrow\mathbb{T}}\right)$.}

\begin{table*}
	\vspace{-0.2cm}
	
	\caption{\label{tab:Ablation-Study} Ablation Study . $\protect\surd$/$\times$
		indicate \emph{using}/\emph{not using} the corresponding step respectively. Numerical results refer to accuracy (\%). }
	
	\begin{centering}
		\hspace{-0in}{\small{}{}}%
		\begin{tabular}{ccccc|cccc|cccc}
			\hline 
			{\small{}{}Dataset }  & \multicolumn{4}{c|}{{\small{}{}AwA}} & \multicolumn{4}{c|}{{\small{}{}aPY}} & \multicolumn{4}{c}{{\small{}{}CUB}}\tabularnewline
			\hline 
			\hline 
			{\small{}{}K-S test }  & {\small{}{}$\surd$ }  & {\small{}{}$\surd$ }  & {\small{}{}$\times$ }  & {\small{}{}$\times$ }  & {\small{}{}$\surd$}  & {\small{}{}$\surd$}  & {\small{}{}$\times$}  & {\small{}{}$\times$}  & {\small{}{}$\surd$}  & {\small{}{}$\surd$}  & {\small{}{}$\times$}  & {\small{}{}$\times$}\tabularnewline
			{\small{}{}Bootstrapping }  & {\small{}{}$\surd$ }  & {\small{}{}$\times$ }  & {\small{}{}$\surd$ }  & {\small{}{}$\times$ }  & {\small{}{}$\surd$ }  & {\small{}{}$\times$}  & {\small{}{}$\surd$}  & {\small{}{}$\times$}  & {\small{}{}$\surd$}  & {\small{}{}$\times$} & {\small{}{}$\surd$}  & {\small{}{}$\times$}\tabularnewline
			\hline 
			{\small{}{}OSL }  & {\small{}{}93.7 }  & {\small{}{}85.6 }  & {\small{}{}37.1 }  & {\small{}{}80.2 }  & {\small{}{}94.3 }  & {\small{}{}85.1}  & {\small{}{}36.8}  & {\small{}{}78.6}  & {\small{}{}59.5}  & {\small{}{}59.3}  & {\small{}{}32.9 }  & {\small{}{}58.6}\tabularnewline
			{\small{}{}G-ZSL }  & {\small{}{}67.3 }  & {\small{}{}63.5 }  & {\small{}{}11.4 }  & {\small{}{}61.7 }  & {\small{}{}58.9 }  & {\small{}{}40.5}  & {\small{}{}6.9}  & {\small{}{}19.5}  & {\small{}{}40.8}  & {\small{}{}38.1}  & {\small{}{}12.1 }  & {\small{}{}31.0}\tabularnewline
			\hline 
		\end{tabular}
		\par\end{centering}
	\begin{centering}
		
		\par\end{centering}
	\vspace{-0cm}
	
\end{table*}

\vspace{0.05in}

\noindent \textbf{Competitors}. We compare several competitors. (1)
\emph{DAP} \cite{lampert2014attribute}, trains a probabilistic attribute
classifier and utilizes the joint probability to predict labels; (2)
\emph{ConSE} \cite{norouzi2013zero}, maps features into the semantic
space by convex combination of attributes; (3) \emph{CMT} \cite{socher2013zero},
projects features into unsupervised semantic space and uses LOF to
detect novel classes; (4) \emph{SSE} \cite{zhang2015zero}, regards
novel classes as mixtures of source proportions to measure the instance
similarity. (5) \emph{Latem} \cite{xian2016latent}, is a novel latent
embedding for ZSL and G-ZSL. (6) \emph{ALE} \cite{akata2016label},
embeds labels into the attribute space by learning a function to rank
the likelihood of each class. (7) \emph{DeViSE} \cite{frome2013devise},
uses both unsupervised information and annotated attributes to classify
classes in an embedding model; (8) \emph{SJE} \cite{akata2015evaluation}
is a hierarchical embedding to learn an inner product gram matrix
between features and attributes. (9) \emph{ESZSL} \cite{romera2015embarrassingly},
focuses on the regularization term in the projection from features
to semantic space. (10) \emph{SYNC} \cite{changpinyo2016synthesized},
aligns the semantic space to feature space by manifold learning. (11)
\emph{SS-VOC} \cite{fu2016semi}, optimizes the triplet loss to learn
the projection from features to semantic space. (12) \emph{SAE} \cite{kodirov2017semantic}
is an auto-encoder to combine feature and semantic space. (13-15)
\emph{PTMCA} \emph{\& SE-GZSL} \& \emph{CADA-VAE }\cite{long2018pseudo,schonfeld2018generalized,verma2018generalized}
leverages VAE \cite{auto_vae} as the generator of pseudo instances
to train the mapping. (16-18) \emph{SP-AEN \& cycle-CLSWGAN \&} \emph{f-CLSWGAN
}\cite{chen2018zero,felix2018multi,xian2018feature} use \emph{GAN}
\cite{wgan} to reconstruct features to balance the target space.
(19) \emph{CDL} \cite{jiang2018learning} aligns semantic and feature
space with dictionary learning.

\vspace{0.05in}

\noindent \textbf{Results}. We use SVM/D-M and SVM/A-G to indicate
the recognition models in source and target/uncertain spaces. As in
Tab.~\ref{tab:GZSL}, our harmonic mean results are significantly
better than all the competitors on almost all datasets. This shows
that ours can effectively address G-ZSL tasks. Particularly -- (1)
Our SVM/D-M results can outperform other competitors over a large
margin on AwA and aPY dataset, due to the efficacy of our compositional
space learning algorithm. Further, with the nonlinear embedding model
-- A-G , our SVM/A-G results are even better on both CUB and AwA
dataset. We argue that the key advantage of our framework comes from
the recognition in the compositional spaces. (3) The good results
of both SVM/D-M and SVM/A-G indicate that our framework is a general
framework. In other words, those previous recognition models are orthogonal
and potentially be useful in each learned compositional space.

Our framework is also applied to a large-scale dataset as Tab.~\ref{tab:G-ZSL-Imagenet}.We
compare several state-of-the-art methods that address G-ZSL on the
large-scale dataset. We use the SVM with the linear projection on
this dataset, due to the huge computational cost on the large-scale
dataset. Our harmonic mean results surpass the other competitors with
a very significant margin. We notice that other algorithms have very
poor performances on $\mathbb{U}\rightarrow\mathbb{T}$. This indicates
the intrinsic difficulty of G-ZSL on large-scale datasets. In contrast,
our algorithm can better separate the testing instances into different
spaces, achieving better recognition performance. Additionally, we
found that the prediction of ConSE \cite{norouzi2013zero} is heavily
biased towards source classes which is consistent with the results
in small datasets. This is due to the probability of target classes
are expressed as the convex combination of source classes.

\subsection{Ablation study}

\noindent \textbf{Open-Set Learning.} Our framework is also evaluated
on the task of OSL. Critically,\textcolor{black}{{} we compare against
	the competitors, including Attribute Baseline (Attrb), W-SVM \cite{scheirer2014probability},
	One-class SVM }\cite{ocsvm}\textcolor{black}{, Binary SVM, OSDN \cite{Bendale_2016_CVPR}
	and LOF} \cite{breunig2000lof}.\textcolor{black}{{} The attribute
	baseline is the variant of our task without using }compositional space
learning\textcolor{black}{{} algorithm. Particularly, the Attrb uses
	the same semantic space and embedding as our model without using the
}compositional space learning\textcolor{black}{{} step, }\textcolor{black}{\emph{i.e.}}\textcolor{black}{,
	using negative samples and prototypes to identify projected instances
	directly. F1-measure is used here as the harmonic mean of source class
	accuracy (specific class) and target prediction accuracy (unnecessary
	to predict the specific class). We summarize the results in Tab.~\ref{tab:Comparison-of-open}.{}
	The accuracy here denotes open class detection accuracy (\%), which is $\frac{N(correct\ classified\ samples)}{N(test\ samples)}$.
	Significant performance gain over existing approaches has been observed,
	in particular for AwA, aPY, and ImageNet. This validates the effectiveness
	of our framework. We attribute the improvement to the newly introduced
	uncertain space which helps better differentiate whether testing instances
	derive from }source\textcolor{black}{{} or target space.}\textbf{ }

\begin{table}[H]
	\centering{}\caption{\textbf{Comparison} of \textcolor{black}{Open-Set recognition algorithms\label{tab:Comparison-of-open}}}
	\begin{tabular}{l|cccc}
		\hline 
		\textcolor{black}{\small{}{}Method / Accuracy}{\small{} } & \textcolor{black}{\small{}{}AwA}{\small{} } & \textcolor{black}{\small{}{}CUB}{\small{} } & \textcolor{black}{\small{}{}aPY}{\small{} } & {\small{}{}ImageNet}\tabularnewline
		\hline 
		{\small{}{}}\textcolor{black}{\small{}{}Attrb}{\small{} } & {\small{}{}33.8 } & {\small{}{}18.7 } & {\small{}{}5.1 } & {\small{}{}3.7}\tabularnewline
		{\small{}{}Binary SVM } & {\small{}{}57.7 } & {\small{}{}29.8 } & {\small{}{}66.6 } & {\small{}{}24.6}\tabularnewline
		\textcolor{black}{\small{}{}W-SVM}{\small{} } & {\small{}{}80.2 } & {\small{}{}58.6 } & {\small{}{}78.6 } & {\small{}{}50.1}\tabularnewline
		{\small{}{}One-Class SVM } & {\small{}{}58.9 } & {\small{}{}27.6 } & {\small{}{}57.1 } & {\small{}{}23.4}\tabularnewline
		{\small{}{}OSDN }\textcolor{black}{\small{}{}}{\small{} } & {\small{}{}49.9 } & {\small{}{}36.7 } & {\small{}{}41.5 } & {\small{}{}--}\tabularnewline
		{\small{}{}LOF } & {\small{}{}60.0 } & {\small{}{}54.5 } & {\small{}{}49.1 } & {\small{}{}38.0}\tabularnewline
		\hline 
		\textcolor{black}{\small{}{}Ours}{\small{} } & \textbf{\textcolor{black}{\small{}{}93.7}}{\small{} } & \textbf{\textcolor{black}{\small{}{}59.5}}{\small{} } & \textbf{\textcolor{black}{\small{}{}94.3}}{\small{} } & \textbf{\small{}{}67.6}\tabularnewline
		\hline 
	\end{tabular}
\end{table}

\vspace{0.05in}

\noindent \textbf{Importance of model selection by bootstrapping}.
We introduce a \emph{variant A} (K-S test ($\surd$) and Bootstrapping
($\times$)) by replacing bootstrapping step as in Sec. \ref{subsec:Determining-the-Initial}
and using Eq (\ref{eq:multiply}) and Eq (\ref{eq:threshold}) to
fix the threshold (\emph{i.e.}, W-SVM \cite{scheirer2014probability}).
As in Tab.\textcolor{black}{~\ref{tab:Ablation-Study}, the} results
of \emph{variant A} are significantly lower than ours on all datasets.
This validates the importance of determining the initial threshold
by bootstrapping.

\vspace{0.05in}

\noindent \textbf{Importance of K-S test. }We define \emph{variant
	B} (K-S test ($\times$) and Bootstrapping ($\surd$)) as the step
without using K-S Test, and compare the results in Ta\textcolor{black}{b.~\ref{tab:Ablation-Study}}.\textcolor{black}{{}
	In particular, we note that }\emph{variant B }has significant lower
results on OSL and G-ZSL than \emph{variant A }and our framework.
One reason is that our bootstrapping step actually learns to determine
a very wide boundary of the source space, to make sure the good results
in labeling testing instances as target space samples (as illustrated
in Fig. \ref{fig:Domain-Division}). The K-S test will further split
the initial source space into source/uncertain space by shrinking the
threshold. Without such a fine-tuning step, \emph{variant B }may wrongly
categorize many instances from target classes as one of the source
classes. Thus, we can show that the two steps of our framework are
very complementary to each other. They work as a whole to enable good
performance on OSL and G-ZSL. Finally, we introduce the \emph{variant
	C} (K-S test ($\times$), and Bootstrapping ($\times$))\emph{ }in
Tab. \textcolor{black}{\ref{tab:Ablation-Study}}\emph{, }by using
W-SVM to do OSL, and then use our ZSL model for G-ZSL. The performance
of \emph{variant C} is again significantly lower than that of ours,
and this demonstrates the efficacy of our model.

\begin{figure}
	\includegraphics[width=1\columnwidth]{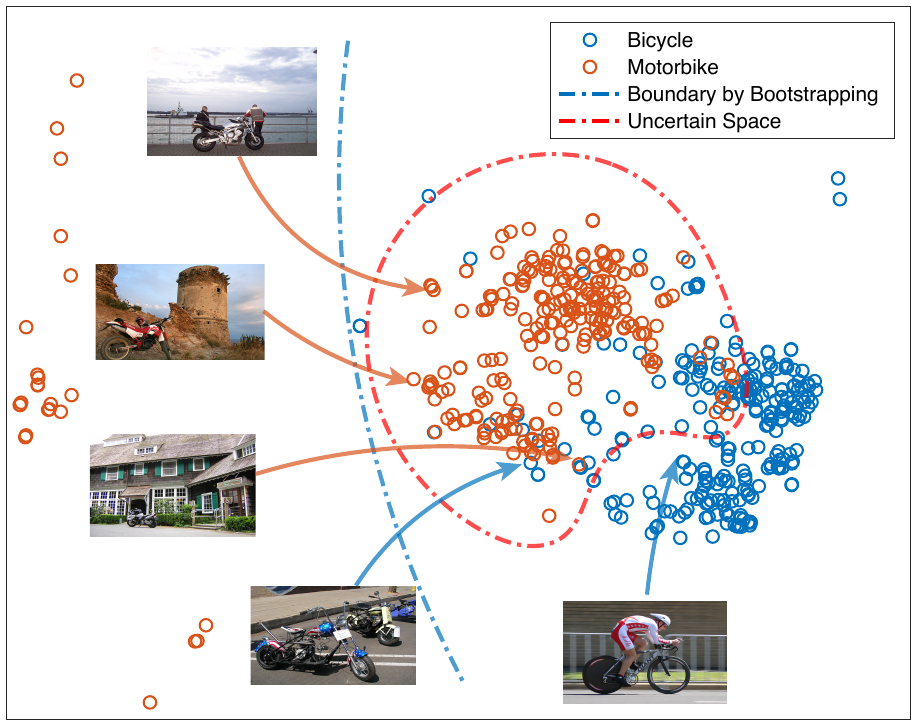}
	
	\caption{t-SNE visualization of Uncertain Space.\label{fig:t-SNE-visualization-of}}
\end{figure}

\noindent \textbf{Visualization of each space}. We use the t-SNE visualization
\cite{tsne} as Fig. \ref{fig:t-SNE-visualization-of} to show each
learned space. Critically, the \emph{bicycle} and \emph{motorbike,}
are one of the source and the target classes in aPY dataset respectively.
The testing instances of motorbike can easily be categorized as one
of source classes (in the initial boundary estimated by bootstrapping),
due to the visual similarity to bicycle. With our K-S test, the instances
of motorbike are labeled into uncertain space and finally correctly
classified by our framework.

\section{Conclusion}

This paper proposes a method that learns to divide the instances into
\emph{source, target and uncertain} spaces for the recognition tasks
from a probabilistic perspective. The compositional space procedure
consists of bootstrapping and K-S Test steps. We use the bootstrapping
to set an initial threshold for each class in the source space. The
K-S test is further employed to fine-tune the boundary between spaces.
Our proposed framework is validated for G-ZSL tasks over many benchmark
datasets and achieves notable results.

\bibliographystyle{abbrv}
\bibliography{iccv2019_conference}

\end{document}